\documentclass[10pt,twocolumn,letterpaper]{article}

\usepackage[pagenumbers]{cvpr} 

\usepackage{graphicx}
\usepackage{amsmath}
\usepackage{amssymb}
\usepackage{booktabs}
\usepackage{siunitx}
\usepackage{comment}
\usepackage{enumitem}
\usepackage{wrapfig}
\usepackage{url}

\usepackage[pagebackref,breaklinks,colorlinks]{hyperref}

\usepackage[capitalize]{cleveref}
\crefname{section}{Sec.}{Secs.}
\Crefname{section}{Section}{Sections}
\Crefname{table}{Table}{Tables}
\crefname{table}{Tab.}{Tabs.}

\usepackage{color}

\DeclareMathOperator*{\argmin}{arg\,min}

\begin{document}

\title{$\boldsymbol{\phi}$-SfT: Shape-from-Template  with a Physics-Based Deformation Model} 

\author{Navami Kairanda\textsuperscript{1, 2}$\;$
Edith Tretschk\textsuperscript{1} $\;$
Mohamed Elgharib\textsuperscript{1}$\;$
Christian Theobalt\textsuperscript{1}$\;$
Vladislav Golyanik\textsuperscript{1}\\
\textsuperscript{1}Max Planck Institute for Informatics, SIC \ \ \ \ \ \ \ \ \textsuperscript{2}Saarland University, SIC
}

\maketitle 

\begin{abstract} 
   Shape-from-Template (SfT) methods estimate 3D surface deformations 
   from a single monocular RGB camera while assuming a 3D state known in advance (a template). 
   This is an important yet challenging problem due to the under-constrained nature of the monocular setting. 
   Existing SfT techniques predominantly use geometric and simplified deformation models, 
   which often limits their reconstruction abilities. 
   In contrast to previous works, this paper proposes a new SfT  approach explaining 2D observations through physical simulations accounting for forces and material properties. 
   Our differentiable physics simulator regularises the surface evolution 
   and optimises
   the material elastic properties such as   
   bending coefficients, stretching stiffness and density. 
   We use a differentiable renderer to minimise the dense reprojection error between the estimated 3D states and the input images and recover the deformation parameters using an adaptive gradient-based optimisation. 
   For the evaluation, we record with an RGB-D camera challenging real surfaces exposed to physical forces with various material properties and textures. 
   Our approach 
   significantly reduces the 3D reconstruction error 
   compared to multiple competing methods. 
   For the source code and data, see  \url{https://4dqv.mpi-inf.mpg.de/phi-SfT/}. 
\end{abstract}

\section{Introduction}

Reconstructing general deformable, temporally-coherent surfaces in 3D from monocular RGB videos is a long-standing, challenging and ill-posed problem. 
It was studied under different assumptions, and methods addressing it can be roughly classified into (template-free) \emph{non-rigid structure from motion}  (NRSfM) \cite{Bregler2000, Garg2013}, (template-based) \emph{shape-from-template} (SfT) \cite{Perriollat2011, Ngo2015}, and neural 3D mesh regression \cite{VMR2020}. 
The objective of SfT %
is: Given a known initial 3D state (a template) of an observed deformable scene or an object, reconstruct all its 3D states observed in the entire image sequence \cite{Salzmann2007}. 
Recent learning-based SfT methods 
encode prior knowledge 
in neural network weights \cite{Shimada2019, FuentesJimenez2021}. 
This offers multiple advantages over a vast body of previous, %
model-based works \cite{Salzmann2007, Salzmann2008, Perriollat2011, Ostlund2012, Yu2015, Ngo2015, Parashar2015}, such as the ability to handle larger deformations, a broader spectrum of supported types of motions and deformations (including highly nonlinear ones), and real-time operation. 
\begin{figure}[t] 
    \begin{center} 
    \includegraphics[width=\columnwidth]{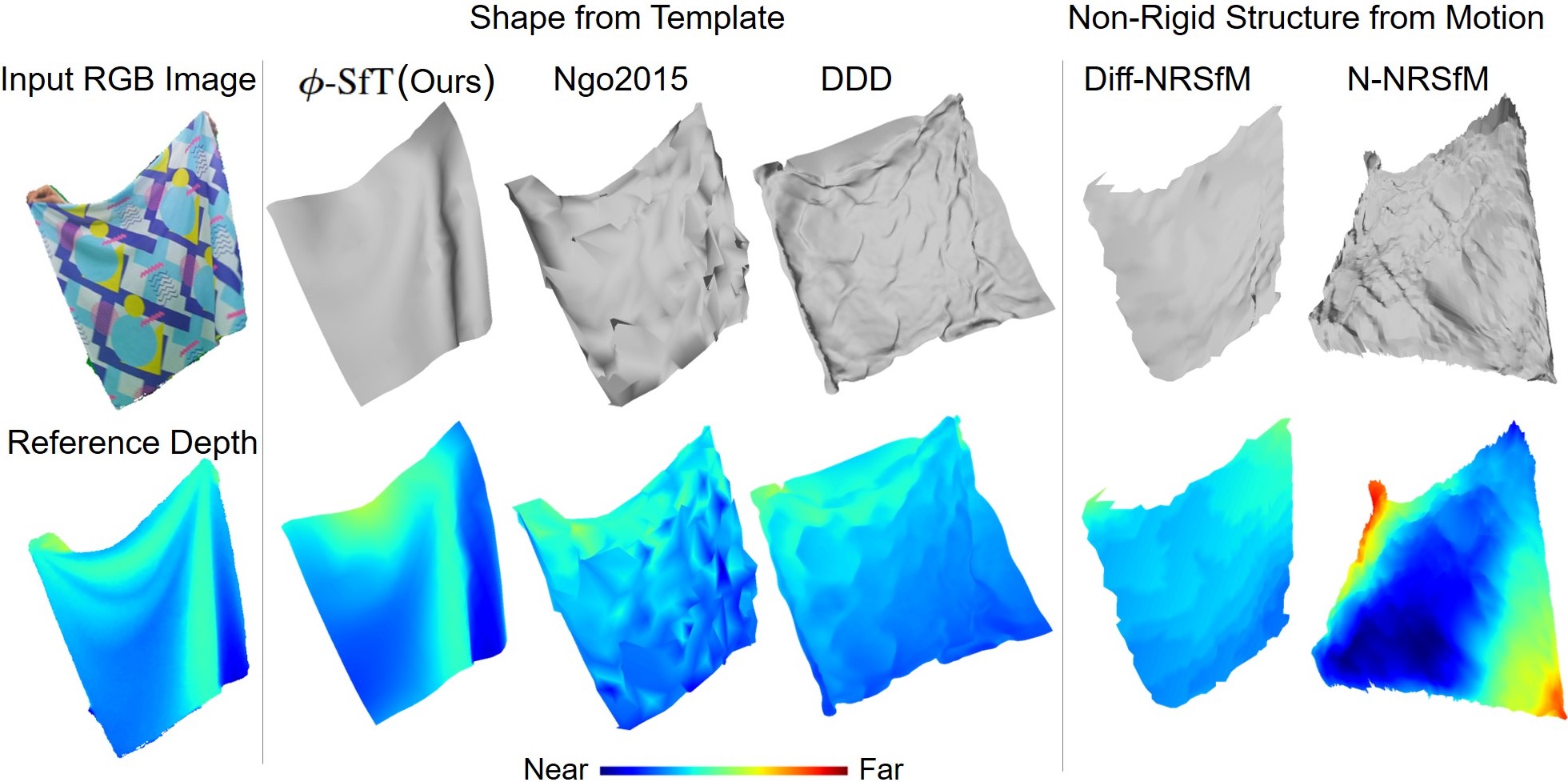} 
    \end{center}
    \vspace{-3pt} 
    \caption{\textbf{Our $\boldsymbol{\phi}$-SfT approach uses a physics simulator to reconstruct challenging deforming 3D surfaces observed in a monocular RGB video.} 
    Compared to %
    existing methods, our estimates are significantly more accurate and physically plausible.} 
    \label{fig:teaser} 
\end{figure}
One of the pivotal limitations of both classical and neural SfT methods is that they capture general 3D states well but not fine local surface deformations. 
This is a consequence of the non-awareness of the physical fold formation process attributable to the elastic properties of the materials and forces acting on them. 
As a result, existing methods can only reconstruct predominantly global deformations.

This paper proposes $\boldsymbol{\phi}$-SfT (from Greek ${\boldsymbol{\phi}}{\upsilon}{\sigma}{\iota}{\kappa}{\eta}$ meaning \textit{physics}): A new analysis-by-synthesis SfT method which addresses several limitations of the current state of the art and improves the accuracy of monocular non-rigid 3D reconstruction by a significant margin; see  Fig.~\ref{fig:teaser} for an overview. 
Our approach explicitly models the physical fold formation process, and its parameters are physically meaningful. %
$\boldsymbol{\phi}$-SfT 
does not require training data. 
We enable gradient-based optimisation by employing two components: 
A differentiable renderer and a differentiable physics simulator. 
Our core idea is to use %
the latter 
as a regulariser during the optimisation of our objective function.
Next, the differentiable renderer ensures that the reprojections of the recovered 3D states accurately match the observed images. 
In contrast to earlier photometric terms used for SfT \cite{Yu2015}, 
differentiable rendering allows us for the first time to define the reprojection error densely \emph{per pixel} and not only \emph{per vertex}. 
We can thus exploit the information present in the texture regardless of the mesh resolution. 
$\boldsymbol{\phi}$-SfT is significantly more accurate than related methods and supports finer-scale local folds, which is shown on a wide spectrum of deformations in extensive experiments (Sec.~\ref{sec:experiments}). 
In summary, this paper has the following technical contributions: 
\begin{itemize}[leftmargin=*]\itemsep0em
   \item A new optimisation-based SfT approach with a physics-based deformation model that ensures high physical fidelity and realism for the surface evolution, outputting temporally smooth 3D shape sequences that are aware of the forces acting on the object (Sec.~\ref{sec:approach}).
   \item Differentiable rendering for SfT to encourage the 2D projections of the reconstructed 3D structure to match the observed images. The differentiability allows optimising for the deformation parameters by minimising a dense per-pixel photometric energy (Sec.~\ref{ssec:objective_function}).
   \item A new dataset of real deforming surfaces recorded in a way to facilitate the quantitative evaluation of reconstruction methods against reference depth maps (Sec.~\ref{ssec:datasets}). 
   The dataset contains surfaces of various textures and  materials, exposed to different external forces. 
   We release our dataset and source code to encourage future research. 
\end{itemize}

\section{Related Work} 

The methods for monocular non-rigid 3D reconstruction 
differ in the assumptions they make about the input, available prior knowledge and how they model the deformations. 
This section reviews methods that can be classified into  non-rigid structure from motion (NRSfM), shape from template (SfT) and monocular 3D mesh reconstruction.

\noindent\textbf{NRSfM} operates on 2D point tracks over the input monocular  views. 
Earlier NRSfM methods were designed for sparse measurements and modelled  deformations with linear subspaces along with  various priors  \cite{Bregler2000, Torresani2008, Akhter2008, Paladini2009, Dai2014}. 
More recent techniques \cite{Garg2013, Ansari2017, Kumar2018, Golyanik_2019} allow reconstructing dense image points observed in a reference frame. 
They impose constraints on spatial point locations to infer smooth and  continuous deforming surfaces. 
A sparse NRSfM method relying on principles of continuum mechanics represents a deformable object using an estimated (not simulated) elastic model and a low-rank force field acting on it \cite{AgudoMoreno2015}. 
Even though the force prior has a direct physical interpretation, this model  still shares most limitations with other methods. 
Diff-NRSfM \cite{Parashar_2020_CVPR} assumes the observed structure preserves its differentiable structure and infinitesimal planarity. 
This method produces impressive results for smooth surfaces but struggles to reconstruct fine-scale folds, unlike our $\boldsymbol{\phi}$-SfT. %

Recently, neural NRSfM approaches both for sparse \cite{Novotny2019, Wang2021} and dense \cite{Sahasrabudhe2019, Sidhu2020} cases were proposed in the literature. 
Some of them need to be trained for each object category \cite{Novotny2019, Sahasrabudhe2019}, whereas N-NRSfM \cite{Sidhu2020} and PAUL \cite{Wang2021} run on unknown data. 
Some 2D keypoint lifting approaches for 3D human pose estimation, such as Chen \textit{et al.}~\cite{Chen2019}, require only 2D data for supervision and share similarities with neural sparse NRSfM. %

\noindent\textbf{SfT} algorithms operate directly on images and assume a known 3D surface prior as input \cite{Salzmann2007, Perriollat2011, Ngo2015, Yu2015}. 
These methods minimise the 3D-2D reprojection error and impose %
geometric constraints such as surface inextensibility \cite{Salzmann2007, Perriollat2011} or isometry \cite{Bartoli2015, Yu2015, Ngo2015}. 
Recent neural SfT methods \cite{Golyanik2018, Pumarola2018, Shimada2019, FuentesJimenez2021} predict 3D surfaces from monocular images relying on datasets with different template states. 
Our $\boldsymbol{\phi}$-SfT contrasts with other SfT methods in that it uses temporal information and a differentiable physics simulator as a regulariser for high-fidelity 3D surface tracking instead of approximating the underlying physical properties via geometric constraints. 
Moreover, none of these methods uses a \emph{per-pixel} differentiable photometric loss which ensures that 3D estimates accurately reproject into the 2D images.

\noindent\textbf{Physics-based priors} in 3D human performance capture is an emerging field, although there is some early work on it relying on multi-view data~\cite{stoll2010}. 
Rempe \textit{et al.}'s~\cite{Rempe2020} method and PhysCap \cite{PhysCapTOG2020,PhysAwareTOG2021} show that integrating physics laws into an objective for sparse 3D human motion capture improves the accuracy of the 3D estimates. 
The proposed constraints reduce the artefacts arising from the monocular setting (\textit{e.g.,} unnatural jitter and body leaning, foot sliding and foot-floor penetration). 
Several methods for 3D human performance capture include \textit{clothes deformations}, such as %
Guo \textit{et al.}~\cite{Guo2021} and Li \textit{et al.}~\cite{Li2021}. 
The method of Guo \textit{et al.}~operates on point clouds and optimises the states of the simulated clothes so that they match the inputs. 
The cloth motion is expressed through a combination of skin friction, gravity and forces attributed to the material (elasticity). 
Thus, their focus is cloth state recovery from sparse point cloud measurements, which provide a strong 3D shape cues, whereas we assume a single 3D template and operate on monocular videos; this inverse problem is much more ill-posed. 
Li \textit{et al.}~generate training data with a physics-based simulator on-the-fly and use it to train a neural network for 3D human performance capture, including clothes deformations. 
Thus, they do not impose hard physics-based constraints as we do with the differentiable physics simulator. 
Liang~\etal~\cite{Liang2019} use 3D supervision for physics-based cloth simulation. 
The work by Weiss~\emph{et~al.}~\cite{Weiss2020CorrespondenceFreeMR} recovers material parameters of a physics simulator in an analysis-by-synthesis policy to solve an inverse elasticity problem. 
In contrast to our method, they additionally require depth inputs for a strong 3D cue and 
do not recover local surface deformations.
The idea of combining a differentiable physics and a differentiable graphics engine has been recently explored in contexts different from ours, \textit{i.e.,} estimation of material properties and visuomotor control  ~\cite{Jaques2020Physics-as-Inverse-Graphics:, murthy2021gradsim, kandukuri2020_diffphys}.

\noindent\textbf{Monocular 3D mesh reconstruction} approaches can be trained on extensive collections of unstructured views in the desired object category. 
Some works~\cite{Choy2016,Wang2018Pixel2Mesh,Bednarik_2021_ICCV} require 3D supervision, while others, similar to ours, do not:
In an early work, Cashman~\etal~\cite{Cashman2013} show that almost-rigid object categories, like dolphins, can be reconstructed from image collections. 
Kanazawa~\etal~\cite{Kanazawa2018} relax the need for input annotations. 
Li~\etal~\cite{VMR2020} extend \cite{Kanazawa2018} to video input and estimate a temporally consistent coarse mesh reconstruction for weakly articulated objects. 
LASR~\cite{Yang2021} further relaxes the need for an initial coarse template. 
We differ from these by the usage of a physics-based deformation model, and we focus on recovering local surface deformations.

\section{Our Approach}\label{sec:approach} 

\begin{figure*}[t]
    \centering
    \includegraphics[width=1\linewidth]{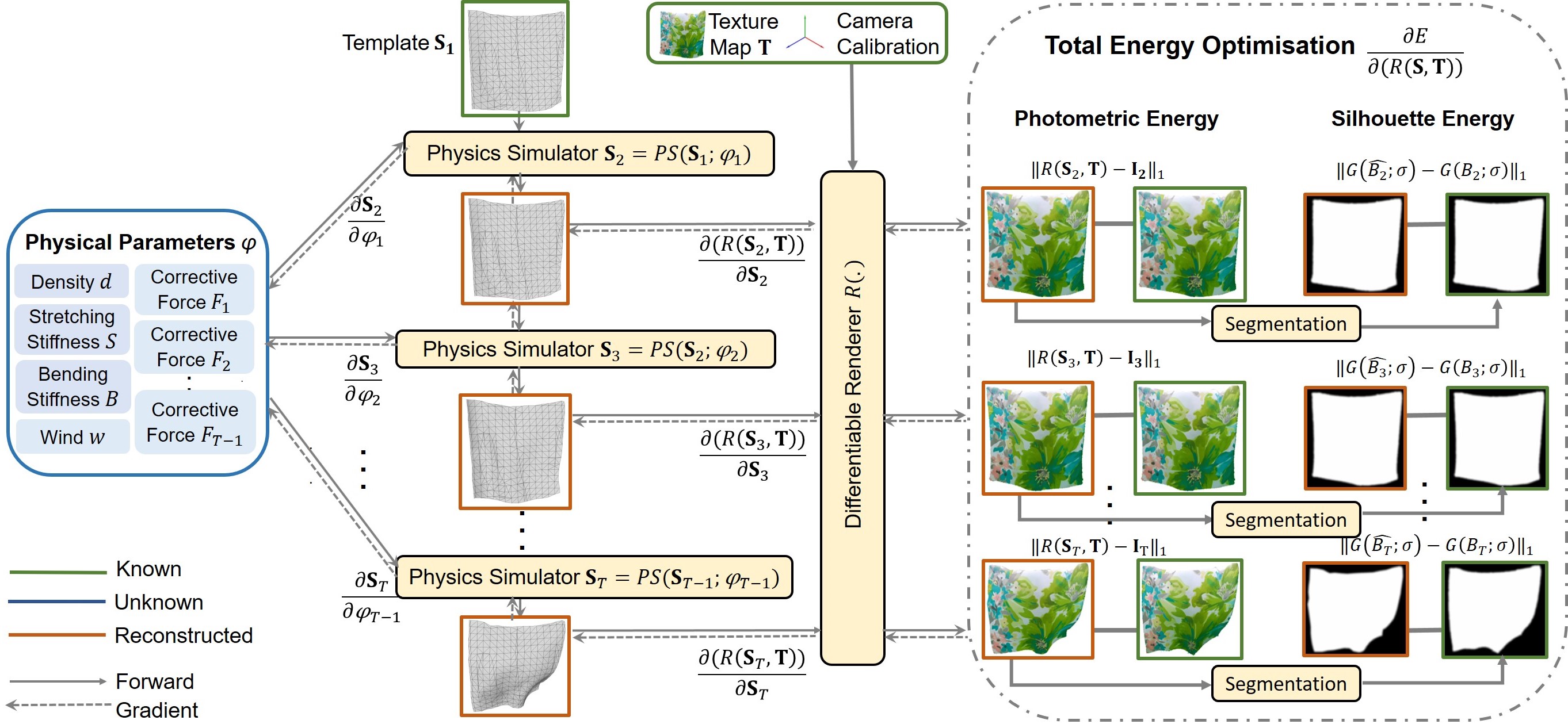}
    \caption{\small Given a sequence of monocular input images $\{\mathbf{I}_t\}_t$, a template at the rest position $\mathbf{S}_1$ and the corresponding texture map $\mathbf{T}$, our technique solves for the unknown physical parameters $\phi$ that describe the deforming 3D surface $\{\mathbf{S}_t\}_t$. 
    We optimise for the per-sequence physical parameters of $\{d, \mathcal{S}, \mathcal{B}, w\}$ as well as the per-frame corrective forces $\{\mathcal{F}_t\}_t$ in a gradient-based manner. 
    We utilise (1) a physics-based differentiable simulator $\mathit{PS}$ for reconstructing meshes with a physical deformation model and (2) a differentiable renderer $R$ for projecting the reconstructions into image space, which allows us to define a reprojection error \emph{over all pixels} (instead of vertices) during optimisation. 
    The differentiable nature of both components enables us to back-propagate the gradients of the total energy $E$ all the way back to the unknown physics parameters. 
    Note that the gradients are calculated automatically and provided here 
    for completeness.
    }

    \label{fig:pipeline}
\end{figure*}

We propose $\boldsymbol{\phi}$-SfT, a new method for the 3D reconstruction of a deforming surface (such as cloth) from a monocular RGB video $\{\mathbf{I}_t\}_{t\in [1,\ldots,T]}$ with known intrinsics. 
As is common for SfT methods \cite{Ngo2015, Yu2015}, we assume that the camera is static and take as input a flat rest shape of the target deformable surface $\mathbf{S}_1$ for $t=1$ with a corresponding texture map $\mathbf{T}$. 
We also assume that a segmentation mask separating a foreground object and background is available. 
To encourage physically plausible deformations, we use a \textit{full} physical model, described in Sec.~\ref{ssec:simulator}, that explicitly models forces acting on the surface and the underlying elastic properties of the material. 
Sec.~\ref{ssec:objective_function} presents the objective function we employ to relate the 2D observations to the estimated reconstructions. 
We then describe how we optimise the objective function for the physical parameters in Sec.~\ref{ssec:optimization} and provide  implementation details in  Sec.~\ref{ssec:implementation}. 
See Fig.~\ref{fig:pipeline} for an overview of
$\boldsymbol{\phi}$-SfT. 

Physics-based simulators are widely used in computer graphics for 3D simulations \cite{10.1145/280814.280821, Narain:2012:AAR}, and differentiable versions exist \cite{Liang2019}. 
Our idea is to use a differentiable physics simulator as a regulariser (deformation model prior) in monocular non-rigid 3D reconstruction. 
Its usage in SfT is, unfortunately, not straightforward.
To integrate the physics-based simulator into our framework, we have to make several crucial improvements to it. 
First, the idealised assumptions of simulated environments cannot account for the variety of forces and effects causing surface deformations in the real world (\textit{e.g.,} wind turbulence). 
We take inspiration from the recent work on physically plausible 3D human motion capture \cite{PhysCapTOG2020}. 
This approach uses a virtual force acting on the root joint of a human skeleton to account for the effects the physics model does not consider. 
We, therefore, introduce corrective forces accounting for mismatched assumptions about the natural scene (Sec.~\ref{ssec:simulator}). 
Second, while most simulators used in computer graphics are designed to create simulations following 3D reference motions, it now has to be driven by the 2D input images, and the gradients need to be backpropagated from the image-based losses. 
Hence, our energy function includes a dense photometric loss and a silhouette loss (Sec.~\ref{ssec:objective_function}). 
Lastly, the optimisation strategy suitable for 3D simulations is not the best choice for our analysis-by-synthesis $\boldsymbol{\phi}$-SfT approach---optimising for deformed surface states, material properties and forces---and we propose an adaptive training strategy instead (Sec.~\ref{ssec:optimization}).

\subsection{Deformation Model}
\label{ssec:simulator}
We seek to reconstruct a sequence of deforming surfaces as 2D manifold meshes in 3D space with fixed topology (edges $\mathbf{E}$), thereby providing temporal correspondences. 
A surface in this sequence can be %
parameterised as a triangular mesh $\mathbf{S}_t = \{ \mathbf{V}_t, \mathbf{E}\}$ where the state of the $i$-th vertex in $ \mathbf{V}_t$ comprises its 3D position $\mathbf{x}^{i}_{t}\in \mathbb{R}^3$ and its velocity $\mathbf{v}^{i}_{t}\in \mathbb{R}^3$.

\paragraph{Surface Parametrisation} 
At the core of our method, we model deformations of the non-rigid surface as a physical process, \textit{i.e.,} as elastic deformations resulting from internal stretching and bending forces as well as external forces acting on the surface. 
We thus do not treat the mesh states $\mathbf{S}_t$ as parameters but instead use a physical parametrisation. 

We initialise the differentiable cloth simulator from the 
template $\mathbf{S}_1$ and generate $\mathbf{S}_t$ with physics simulation $\mathit{PS}$ according to the material parameters and external forces: 
\begin{equation}
    \mathbf{S}_t = \mathit{PS}(\mathbf{S}_{t-1};\phi_{t-1}), 
    \label{eq:deformation_model}
\end{equation} 
where $\phi_{t-1}$ are the estimated physics parameters, \textit{i.e.,}  
\begin{equation}
    \phi_{t-1} = \{d, \mathcal{S}, \mathcal{B}, w, \mathcal{F}_{t-1}\}.
\end{equation}
Here, material density $d\in\mathbb{R}$, stretching stiffness %
$\mathcal{S}\in\mathbb{R}^{24}$ (resistance to stretching), and bending stiffness $\mathcal{B}\in\mathbb{R}^{15}$ (resistance to bending and folding) all together describe the elastic properties of the material and, hence, determine the cloth's internal forces. 
We also optimise for the wind force $w\in\mathbb{R}^3$. %
However, the wind model is not sufficient to fully describe all the external forces in the scene, such as hand contacts and wind turbulence. 
We seek to correct for these model insufficiencies  %
by additionally defining a set of corrective forces $\mathcal{F}=\{\mathcal{F}_t \in \mathbb{R}^{|\mathbf{V}_t|\times3}\}_{t\in[1,\ldots,T-1]}$. 
Note that these vary across vertices and across time. 
They can, in principle, account for any physical force that the simulator does not explicitly model. 
In the following, we use the shorthand $\phi = \{d, \mathcal{S}, \mathcal{B}, w, \mathcal{F} \}$.

\paragraph{Physics Simulator} 
For $\mathit{PS}$, we follow the cloth simulation pipeline introduced by Narain~\etal~\cite{10.1145/280814.280821,10.1145/2461912.2462010, Narain:2012:AAR}. 
In the continuous domain, physics-based simulation can be formulated as a time-varying partial differential equation~\cite{10.1145/280814.280821}: 
\begin{equation}
    \frac{\partial^2 \mathbf{x}}{\partial t} = \mathbf{M}^{-1}\mathbf{f} (\mathbf{x}, \mathbf{v}),  
    \label{eq:diff_sim_problem}
\end{equation} 
where $(\mathbf{x}, \mathbf{v})$ is the vertex state, and $\mathbf{M}$ is a diagonal matrix of the mass distribution derived from the material density $d$ and surface area. 
$\mathbf{f}(\cdot)$ are the forces, \textit{i.e.,} internal forces which are a function of cloth elastic properties $\mathcal{S}$ and $\mathcal{B}$ as well as external forces such as wind $w$ and corrective forces $\mathcal{F}_t$. 
We follow the elastic model for cloth material %
of Wang~\etal~\cite{WangDDE} for describing the effects of $d$, $\mathcal{S}$ and $\mathcal{B}$. 
This model approximates various nonlinear and anisotropic behaviours seen in cloths when subject to external forces. 

In practice, we are given the known position $\mathbf{x}_{t-1}$ and velocity $\mathbf{v}_{t-1}$ of the system at time $t{-}1$. %
Our goal is to determine the new position $\mathbf{x}_t = \mathbf{x}_{t-1} + \Delta{\mathbf{x}}$ and velocity $\mathbf{v}_t = \mathbf{v}_{t-1} + \Delta{\mathbf{v}}$ at time $t$ with a time step size $h{=}1$. 
To that end,  \eqref{eq:diff_sim_problem} can be transformed into a first-order differential equation and 
solved for $\Delta{\mathbf{x}}$ and $\Delta{\mathbf{v}}$ with the implicit, backward Euler method \cite{10.1145/280814.280821}: 
\begin{equation}
    \begin{pmatrix}
        \Delta{\mathbf{x}}  \\
        \Delta{\mathbf{v}} 
    \end{pmatrix}
    = h 
    \begin{pmatrix}
        \mathbf{v}_t \\ 
        \mathbf{M}^{-1} \mathbf{f}(\mathbf{x}_t, \mathbf{v}_t)
    \end{pmatrix},
    \label{eq:diff_sim_solution1}
\end{equation} 
which is nonlinear due to $\mathbf{f}$. 
To turn \eqref{eq:diff_sim_solution1} into a linear system, $\mathbf{f}$ can be approximated via linearisation:
\begin{equation}
    \mathbf{f}(\mathbf{x}_t, \mathbf{v}_t) = \mathbf{f}_{t-1} +  \frac{\partial \mathbf{f}}{\partial \mathbf{x}}  h (\mathbf{v}_{t-1} + \Delta{\mathbf{v}}) + \frac{\partial \mathbf{f}}{\partial \mathbf{v}} \Delta{\mathbf{v}},
    \label{eq:diff_sim_solution2}
\end{equation} 
where the Jacobians $\frac{\partial \mathbf{f}}{\partial \mathbf{x}}$ and $\frac{\partial \mathbf{f}}{\partial \mathbf{v}}$ are evaluated at $\mathbf{f}_{t-1}$. 
Thus, a simple cloth simulation process involves solving for $\mathbf{v}_t$ using \eqref{eq:diff_sim_solution1} and \eqref{eq:diff_sim_solution2}, and computing the subsequent simulation state as $\mathbf{x}_t = \mathbf{x}_{t-1} + h \mathbf{v}_t$.

There can additionally be self-collisions and collisions with a dynamic obstacle mesh $(\mathbf{x}_t^{obs},\mathbf{v}_t^{obs})$ during simulation. 
Harmon~\etal~\cite{RTSC} determine the collision response at the impact zones to update the vertex positions accordingly: %
\begin{equation}
\mathbf{x}_t = \mathbf{x}_{t} + \operatorname{collision\_response} (\mathbf{x}_t, \mathbf{v}_t, \mathbf{x}_t^{obs}, \mathbf{v}_t^{obs}). 
\label{eq:diff_sim_collision}
\end{equation}
Since we want to use end-to-end gradient-based optimisation, %
we need to backpropagate gradients through this extra step. %
However, due to the high dimensionality of the dynamical system when modelling cloth, a na\"ive gradient computation for the general system \eqref{eq:diff_sim_solution1} and 
\eqref{eq:diff_sim_collision} 
(the collision response) can become impractical. 
Liang~\etal~\cite{Liang2019} propose a solution for this problem, and we proceed with their approach. 
Specifically, they use implicit differentiation for \eqref{eq:diff_sim_solution1} and \eqref{eq:diff_sim_collision}, where the gradient of the latter is approximated via QR decomposition of a much smaller constraint matrix. 
For more details on the backward pass,  please refer to \cite{Liang2019}. 
We next describe the objective function we use to optimise for the parameters $\phi$ of the differentiable simulator.

\subsection{Our Objective Function} 
\label{ssec:objective_function} 
We now have a physical deformation model that is parametrised by $\phi$, and that outputs a 3D mesh $\mathbf{S}_t$ for time $t$. 
We solve for the optimal parameters $\phi^*$ by minimising the objective function $E = E_p + \lambda E_s$ (with $\lambda\in\mathbb{R}$): 
\begin{equation}
    \phi^* = \argmin_\phi E(\phi). 
    \label{eq:full_problem}
\end{equation}

Since we are only given RGB images for time $t{>}1$, we define a photometric energy term $E_p$ in the image space. 
Specifically, $E_p$ is an $\ell_1$ RGB data term to encourage photometric consistency between the reconstructed surface rendered into 2D and the input frames:
\begin{equation}
    E_p = \sum_{t=2}^{T} \| R ( \mathbf{S}_t, \mathbf{T} ) - \mathbf{I}_t \|_{1},
\end{equation}
where $R(\cdot)$ is a differentiable renderer outputting a perspective projection of the input mesh (textured with $\mathbf{T}$) onto the image plane with known intrinsics. 
We implement $R$ with \textit{Soft Rasterizer}~\cite{liu2019softras}, which introduces useful gradients by composing probability maps of rendered triangles into the final image. 
This allows us 1) to define $E_p$ densely \emph{over all pixels}, instead of just the vertices;  
2) use the information in the high-resolution %
$\mathbf{T}$
that would have been ignored if we had used a photometric term only at the vertices. %

While the photometric energy term works well for local corrections, it does not provide a signal for mismatches that are farther apart in the image space. 
To get a signal even for larger, coarser errors, we add a silhouette energy term that encourages consistency between the foreground segmentation mask of the input frames and the rendered 2D surface: 
\begin{equation}
\label{eq:smooth}
   E_s = \sum_{t=2}^{T} \| G({B}_t;\sigma) - G(\hat{B}_t;\sigma) \|_{1}.
\end{equation}
Here, $B$ and $\hat{B}$ are the foreground binary segmentation masks of the reconstructed and the input images, respectively. 
$G(\cdot,\sigma)$ represents a Gaussian filter of standard deviation $\sigma$. 
The Gaussian filter smooths the binary masks, extending the spatial area where informative gradients are obtainable. 
Without it, 
non-zero gradients would be obtained only at pixels located immediately next to the silhouettes of both binary masks. 
Thus, if the silhouettes do not match almost perfectly at a pixel, the gradient would be zero there, providing no signal to the network as to the target direction to move the mesh's triangles. 
Importantly, both the ground-truth mask $B$ and the rendered mask $\hat{B}$ are processed in the same way, ensuring that $E_s$ is well-behaved. 

Given our model and the objective function, we next look at how we solve the optimisation problem. 

\subsection{Optimisation} 
\label{ssec:optimization}

Our goal is to obtain the optimal physical parameters $\phi^*$ via \eqref{eq:full_problem}. 
We use iterative, gradient-based optimisation to that end. 
Since both the simulator $\mathit{PS}$ and the renderer $R$ are differentiable, we can back-propagate gradients from 
the objective function 
$E$ through the rendering to 
$\mathbf{S}_t$ and from there further through the physics simulation to the physical parameters $\phi$ and further to all earlier meshes (see Fig.~\ref{fig:pipeline}).

\paragraph{Initialisation}
To obtain a sufficiently accurate initial guess for the elastic properties $d$, $\mathcal{S}$, and $\mathcal{B}$, we set them to the average values of ten different real fabrics \cite{WangDDE}. 
The wind and corrective forces $\mathcal{F}$ are initialised to $\mathbf{0}$, \textit{i.e.,} a zero vector. 
Note that this initialisation leads $\mathit{PS}$ to initially generate surfaces $\{\mathbf{S}_t\}_t$ that are identical to the template $\mathbf{S}_1$. %

\begin{wrapfigure}{l}{0.20\textwidth}
\vspace{-22pt} 
  \begin{center}
    \includegraphics[width=0.23\textwidth]{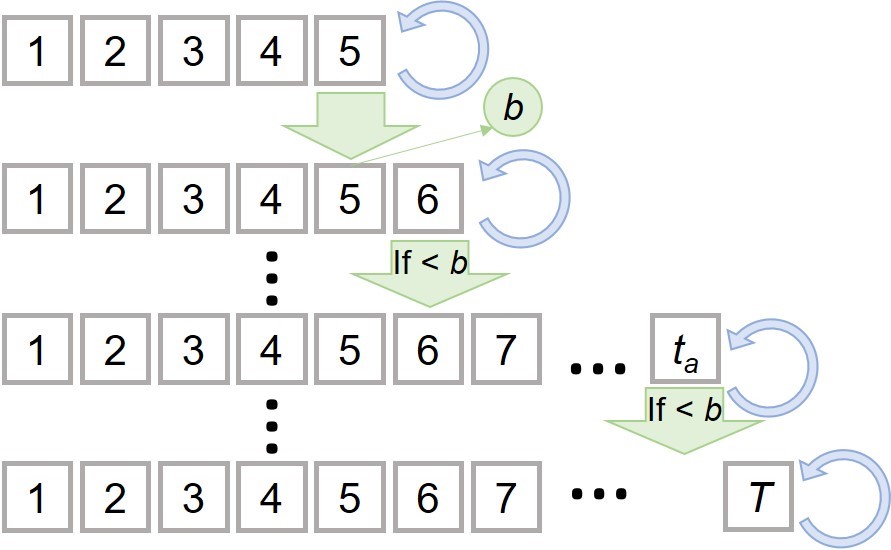}
  \end{center}
  \vspace{-12pt}
  \caption{Our adaptive optimisation scheme.}
  \vspace{-12pt}
  \label{fig:adaptive_scheme} 
  \hspace{-15pt}
\end{wrapfigure}

\paragraph{Adaptive Optimisation Scheme} 
Since simulation is a temporal process, 
$\phi$
for the early frames directly influence the reconstructions of the later frames. 
In addition, 
later frames
are initially reconstructed at lower fidelity than earlier ones. 
Therefore, similar to tracking a surface, we exploit the temporal frame order 
and 
do not optimise for all $t{\geq}1$ from the start. 
Instead, we adaptively grow the  \emph{active} temporal window, 
\textit{i.e.,} the latest time $t_a$ up to which all earlier frames $t\leq t_a$ participate in the optimisation. 
We start with the first five frames as active and optimise $E$ \eqref{eq:full_problem} for them. 
Once the energy of the latest frame decreases below a threshold $b$, we add the next frame to the optimisation; see Fig.~\ref{fig:adaptive_scheme}.
$b$ is set to the energy that the fifth frame has when the sixth frame is added. %
We assume the fifth frame to be well-reconstructed since it is early in the sequence. 
(In case the optimisation cannot reach the threshold, we set a maximum number of iterations, after which we progress regardless.) 
This adaptive scheme speeds up optimisation by converging to a reasonable guess for $\phi$ before later frames become active. 
Moreover, it allocates more iterations to frames with more challenging deformations. 
We evaluate the adaptive optimisation scheme 
experimentally in Sec.~\ref{ssec:ablation}.

\subsection{Implementation  Details}\label{ssec:implementation} 
We implement $\boldsymbol{\phi}$-SfT in  Pytorch~\cite{ravi2020pytorch3d}. 
\eqref{eq:full_problem} is solved with the Adam optimiser~\cite{kingma2017adam} with  learning rate $10^{-3}$. 
The adaptive optimisation scheme leads to several hundred %
iterations in most cases, which takes $16{-}24$ hours on an Nvidia RTX 8000 GPU. 
Due to the sequential nature of the simulator, we compute the objective function for all active times $t$ for each optimisation iteration. 
We set $\sigma{=}7\mathit{px}$, $\lambda{=}0.5$ for $E_s$ and apply the corrective forces $\mathcal{F}$ by modifying the velocity of vertex $i$ at time $t$: $v_t^i = v_t^i + \mathcal{F}_t^i $ (because both mass and time steps  are constant). 
We keep the wind air density fixed at $1$\si{kg/m^3} and optimise only for the wind velocity. 

The images in our real scenes have resolution $1920{\times}1080$ pixels. %
We also pre-process the real scenes (recorded with an RGB-D camera): We first segment out the background from the captured images and point clouds by depth thresholding. 
We next use Poisson 
surface reconstruction~\cite{Kazhdan13}  on the template (at $t=1$), which yields ${\sim}300$ vertices on average. %
We then determine the initial rigid pose relative to a flat sheet (which is required by the simulator) using iterative closest point (ICP)~\cite{besl1992method}, and initialise the simulator with it. 
We obtain the texture map $\mathbf{T}$ by projecting the vertices of the template mesh $\mathbf{S}_1$ onto the image space of the first image $\mathbf{I}_1$ with known camera intrinsics.

\section{Experimental Results}\label{sec:experiments}

We evaluate our technique on real and synthetic data.
We recorded the \textit{$\boldsymbol{\phi}$-SfT real dataset} (Sec.~\ref{ssec:datasets}) 
of natural sequences using a monocular RGB camera and the depth camera Azure Kinect. 
The latter is used to obtain pseudo-ground-truth segmentations and deformations. 
Qualitative and quantitative results on this dataset in Sec.~\ref{ssec:reala_sequences} show our technique clearly outperforms the 
state of the art by capturing a wider variety of deformations and local folds. 
In addition, we generate a new \textit{$\boldsymbol{\phi}$-SfT synthetic dataset} of four monocular RGB sequences of naturalistically deforming surfaces using physics-based simulation; see our supplement for further details and evaluations, likewise demonstrating the superior performance of $\boldsymbol{\phi}$-SfT. 
We also perform an ablation study in Sec.~\ref{ssec:ablation} that demonstrates the importance of corrective forces and other design choices. 

\subsection{The $\boldsymbol{\phi}$-SfT Real Dataset}
\label{ssec:datasets}
\begin{figure}
    \centering
    \includegraphics[width=\columnwidth]{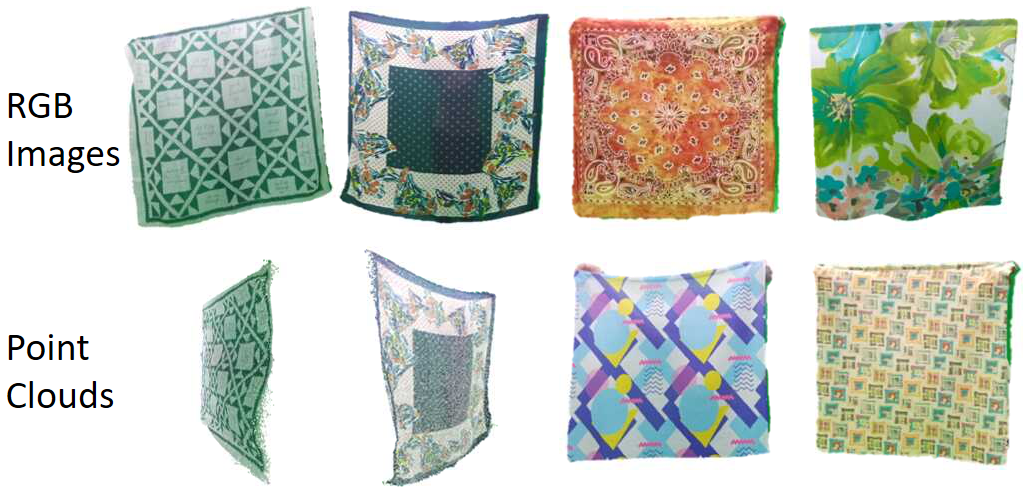}
\caption{We record a new \textit{$\boldsymbol{\phi}$-SfT real dataset} of nine sequences with reference depth data to facilitate quantitative comparisons of monocular 3D surface reconstruction methods. 
Our setup consists of a synchronised RGB camera and an Azure Kinect. 
The depth camera provides depth maps serving as pseudo ground truth. 
}
    \label{fig:dataset}
\end{figure}

We have recorded a new dataset of deforming surfaces to allow quantitative evaluations of reconstruction methods on real data against pseudo ground truth. 
The dataset includes 
nine sequences of various surface shapes and textures, including differing material properties due to  differences in the cloths' fabric and weaving (\textit{e.g.,} there are more and less elastic and more and less dense materials). 
The texture pattern varies from fine-grained and regular to more global and irregular patterns; the cloth size ranges from $55{\times}55\mathit{cm}$ to $95{\times}95\mathit{cm}$. 
The surfaces are exposed to external forces, \textit{i.e.,} gravity, wind, and hand contacts. 
Each sequence is simultaneously recorded using a monocular RGB and depth camera and has a length of about $40$ frames, such that they focus on challenging folds. 
To obtain the texture map $\mathbf{T}$ under the same lighting conditions as the recorded sequence, we start with a flat rest state for $t=1$ and obtain $\mathbf{T}$ from it. 
Then, pseudo-ground-truth deformations for each frame are reconstructed as point clouds through backprojection, using the depth images and known camera intrinsics. 
The point clouds are provided in absolute distance units (meters). 
Fig.~\ref{fig:dataset} shows an overview of the recorded sequences. 

\subsection{Comparison with Existing Methods}
\label{ssec:reala_sequences}

We compare our technique to SfT methods, namely Yu~\etal's Direct, Dense, Deformable (DDD)~\cite{Yu2015}, Ngo~\etal's Ngo2015~\cite{Ngo2015} and Shimada~\etal's IsMo-GAN~\cite{Shimada2019}, as well as %
Sidhu~\etal's Neural NRSfM (N-NRSfM)~\cite{Sidhu2020} and Parashar~\etal's Diff-NRSfM~\cite{Parashar_2020_CVPR}. 

Since NRSfM methods accept 2D point correspondences, we track 2D points densely across the input images with multi-frame subspace flow (MFSF) \cite{Garg2013MFOF, MFSF}, as suggested in \cite{Sidhu2020}. 
The first frame of the sequence is selected as a keyframe for tracking. 
We provide DDD with the required hierarchy of coarse-to-fine templates and Ngo2015 with the same template as ours. 
To demonstrate the necessity for the physical simulation based on the
internal stretching and bending forces,
we show results of the \emph{``Only $\mathcal{F}$''} baseline where the only forces 
are the correctives %
$\{\mathcal{F}_t\}_t$. 
As other forces and parameters are omitted, we implement this baseline as optimisation of per-vertex offsets over time. 

Due to the monocular scale and depth ambiguities, we align reconstructions of all methods to the ground truth in a rigid-body fashion. 
For the first frame, we determine the transformation matrix using Procrustes alignment \cite{umeyama1991least}, which is further refined with rigid ICP~\cite{besl1992method} for later frames. 

\begin{figure*}[t!]
\begin{center}
   \includegraphics[width=1.0\linewidth]{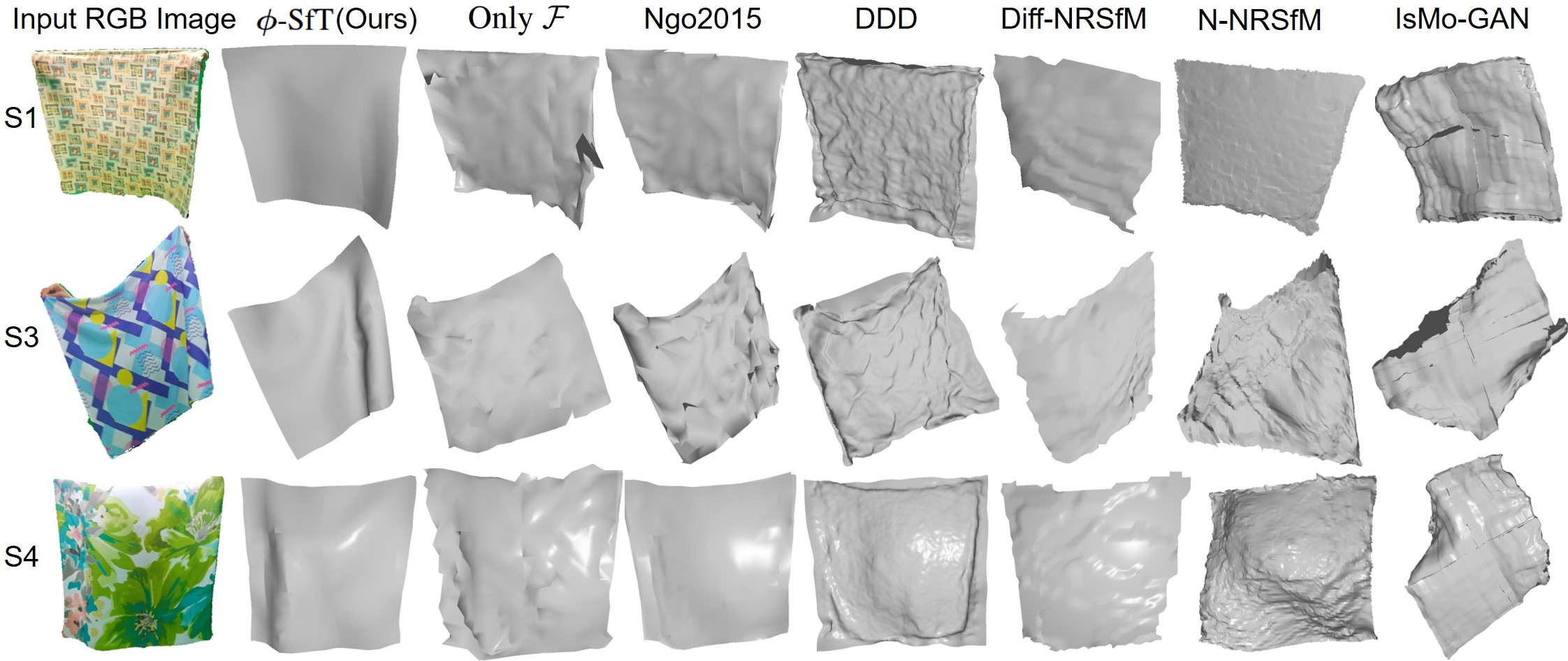}
\end{center}
   \vspace{-8pt} 
   \caption{ 
   Qualitative comparisons of several tested methods \cite{Ngo2015, Parashar_2020_CVPR, Sidhu2020, Yu2015, Shimada2019} and $\boldsymbol{\phi}$-SfT 
   for a representative frame of the %
   S1, S3 and S4 sequences. 
   Our results are significantly more accurate and, unlike the other methods, capture the folds well (especially for S3). 
   }
\label{fig:visual_comparison}
\end{figure*}

For quantitative evaluation, we compute the Chamfer distance between the pseudo-ground-truth point cloud from Kinect $G=\{\mathbf{g}_i\in\mathbb{R}^3\}_i$ and points $M=\{\mathbf{m}_j\in\mathbb{R}^3\}_j$ sampled from the reconstructed mesh: 
{\small
\begin{align}
    \mathit{Ch}_{G,M} = 
    \frac{1}{\lvert G \rvert} \sum_{\mathbf{g}\in G} \operatorname*{min}_{\mathbf{m}\in M} \lVert \mathbf{g} - \mathbf{m}  \rVert^2_2 %
    +\frac{1}{\lvert M \rvert} \sum_{\mathbf{m}\in M} \operatorname*{min}_{\mathbf{g}\in G} \lVert \mathbf{m} - \mathbf{g}  \rVert^2_2. 
\end{align}}
\hspace{-3pt}We report average Chamfer distance per sequence $\tilde{\mathit{Ch}}_{G,M}$. 
Figs.~\ref{fig:teaser} and  \ref{fig:visual_comparison} show that  $\boldsymbol{\phi}$-SfT substantially outperforms all tested methods qualitatively; see the supplement for more  visualisations of our results. 
Tab.~\ref{tab:numerical} shows that $\tilde{\mathit{Ch}}_{G,M}$ of  $\boldsymbol{\phi}$-SfT is on average lower than $\tilde{\mathit{Ch}}_{G,M}$ of competing methods.  %

Our results confirm that SfT and NRSfM, both relying on simple geometric prior assumptions (such as surface smoothness, isometry or small local deformations), cannot cope with the elaborate fold patterns present in our dataset. 
The results of N-NRSfM follow the silhouettes of the input images better than DDD and IsMO-GAN, thanks to 2D point tracking, even though its $\tilde{\mathit{Ch}}_{G,M}$ is the highest (due to 
rigidity assumption for the initialisation 
\cite{TomasiKanade1992, Sidhu2020}). 
Its 3D surfaces are somewhat rugged, and the surface pattern allows to recognise the observed texture in the second row. %
Moreover, %
as expected, more fine-grained textures result in more accurate 2D point tracking by MFSF. 
DDD does not track large deformations and silhouettes well in our tests. 
IsMo-GAN, trained on rather smooth surfaces, cannot reproduce local folds and barely captures the contours.
Diff-NRSfM produces reasonable reconstructions in smooth regions owing to its differentiable structure-preserving formulation. 
However, it is sensitive to noise in 2D measurements, which leads to visual artefacts in the regions with challenging folds. 
Ngo2015 failed on three scenes (partially on S7 and entirely on S8 and S9%
). 
Ngo2015 mostly produces physically implausible results 
but achieves lower $\tilde{\mathit{Ch}}_{G,M}$ on a few scenes; qualitative observations, however, remain the same for all sequences (\cref{fig:teaser,fig:visual_comparison}). 
This suggests that an isometry prior \cite{Ngo2015} is less effective 
compared to our physics-based elastic model, which can even express  non-isometric deformations (depending on the parameters). 
When removing all forces from the model except for the correctives, the results degrade in quality and average error increases ${>}50\%$, see ``Only $\mathcal{F}$'' in Fig.~\ref{fig:visual_comparison} and  Tab.~\ref{tab:numerical}. 
This experiment demonstrates that the accuracy of the full  $\boldsymbol{\phi}$-SfT model does not rely solely on the corrective forces. 
In contrast, $\boldsymbol{\phi}$-SfT estimates temporally coherent surfaces and captures all significant folds while missing only small 
nuances. As seen in Fig.~\ref{fig:teaser}, our physics-based model provides a reasonable prior for self-occluded surface parts. Moreover, our method is not sensitive to initialisation. We empirically find no issue with always initialising with our default material parameters.

\begin{figure}[t!]
\begin{center}
   \includegraphics[width=0.97\linewidth]{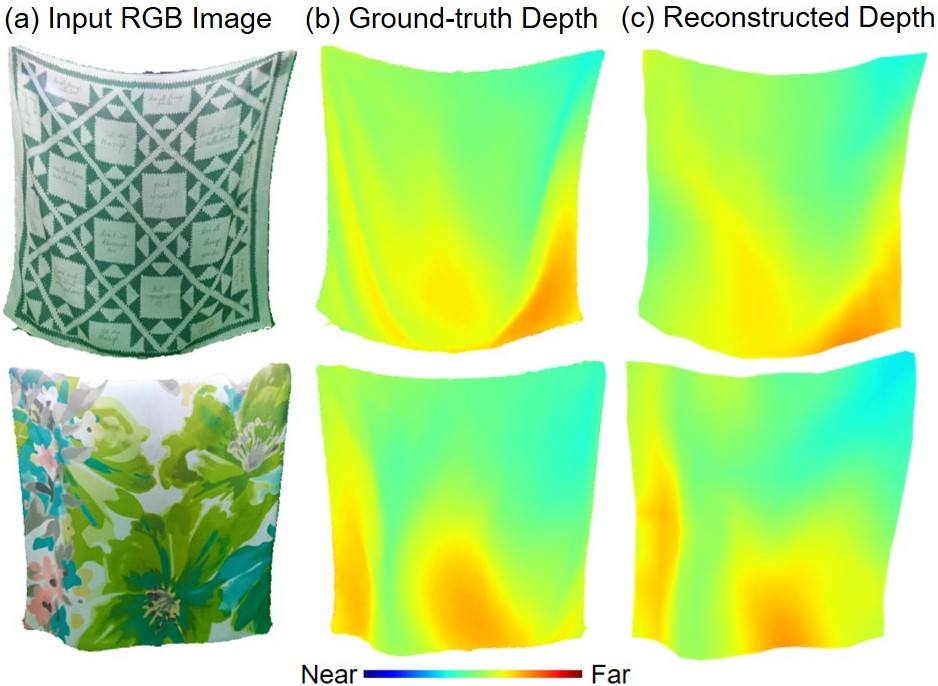} 
\end{center}
    \vspace{-8pt}
   \caption{We show qualitative results as  colour-coded depth maps. 
   For (a) the given RGB image, (b) the ground-truth depth map exhibits similar features as (c) our reconstructed depth. 
   Both the coarse shape and local folds are well 
   captured. 
   }
\label{fig:visual_depthmaps}
\end{figure}

See Fig.~\ref{fig:visual_depthmaps} for depth maps reconstructed by $\boldsymbol{\phi}$-SfT. 
This alternative way to visualise the results allows us to study and perceive even smaller surface details. 
We also strongly encourage the readers to watch our supplementary video.

\newcommand\scale{0.8}

\begin{table}
  \centering 
  \resizebox{\columnwidth}{!}{
  \begin{tabular}{rrrrrrrr} 
    \toprule
    Seq.
    & IsMo-GAN 
    & N-NRSfM 
    & DDD 
    & Diff-NRSfM 
    & Ngo2015 %
    & Only $\mathcal{F}$ %
    & $\boldsymbol{\phi}$-SfT\\
    \midrule
    S1   & 19.69 & 8.25  & 2.95  & 17.14 & 2.19 & 2.59 & \textbf{0.79}\\ 
    S2   & 22.18 & 33.62 & 1.69  & 4.46  & \textbf{1.51} & 1.60 & 2.75  \\ 
    S3   & 33.54 & 104.60& 3.80  & 4.40  & \textbf{2.17} & 3.23 & 3.54 \\ 
    S4   & 90.30 & 77.02 & 25.73 & 41.37 & 15.90 & 14.95 & \textbf{7.60} \\
    S5   & 92.78 & 72.66 & 10.46 & 26.92 & 10.72 & 21.32 & \textbf{6.15}  \\ 
    S6   & 57.62 & 8.73  & 6.97  & 14.02 & \textbf{3.01} & 3.08 &  3.14 \\ 
    S7   & 49.27 & 129.44& 15.64 & 12.49 & 7.95* & 6.03 &\textbf{4.73} \\
    S8   & 24.45 & 38.06 & 7.61  & 9.91  & fail & 3.78 &\textbf{2.52}\\ 
    S9   & 53.12 & 19.81 & 11.77 & 5.29  & fail & 4.39 &\textbf{2.36} \\ 
    \hline 
    Avg.    &  49.22 &  54.69 & 10.87 & 15.11 & 5.92* & 6.77 & \textbf{3.93} \\
    
    \bottomrule
  \end{tabular}
  }
  \caption{
  We quantitatively compare $\boldsymbol{\phi}$-SfT to the state of the art on the real dataset. 
  The average Chamfer distance $\tilde{\mathit{Ch}}_{G,M}$ is multiplied by $10^4$ for readability. 
  ``$^*$'': Ngo2015 failed on the last few frames of S7, which we exclude from the error computation. 
  }
  \label{tab:numerical}
\end{table}

\subsection{Ablative Study}
\label{ssec:ablation}

\begin{figure}[t!]
\begin{center}
   \includegraphics[width=\columnwidth]{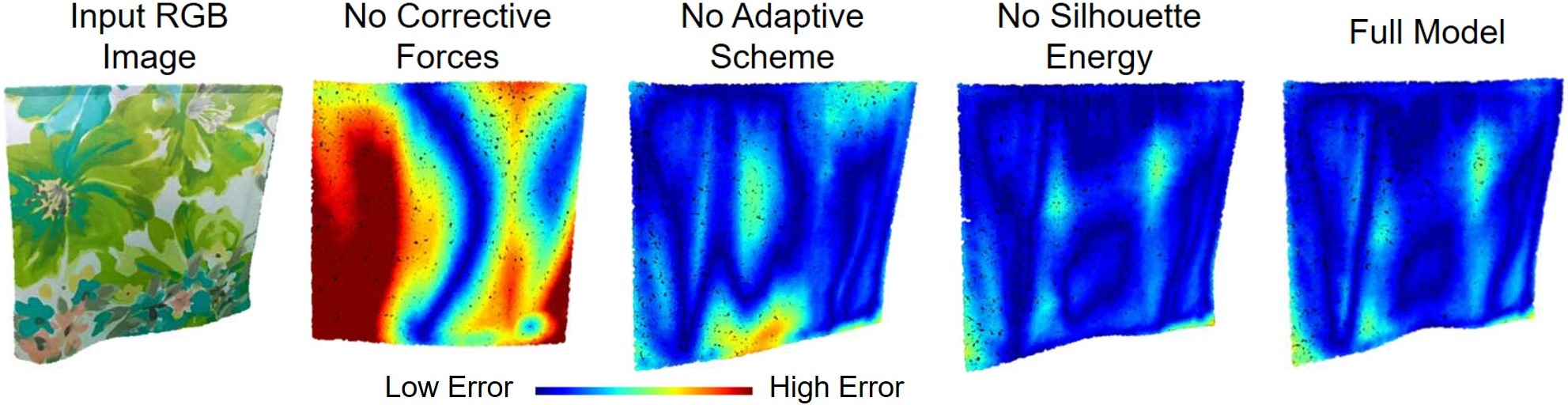} 
\end{center}
   \vspace{-8pt}
   \caption{
   In the ablative study, we remove corrective forces, the adaptive scheme, or the silhouette energy term. The corrective forces are the most crucial component to make our method work. 
  }
\label{fig:ablation}
\end{figure}

We conduct an ablative study on the various design choices %
to integrate the physics simulator into the SfT setting. 
We test the following modes: 1) Operation without corrective forces $\mathcal{F}$, 2) Influence of the adaptive training by considering all frames from the start (Sec.~\ref{ssec:optimization}), and 3) Disabling the silhouette term \eqref{eq:smooth}. 
On average across all sequences, we obtain 
$\tilde{\mathit{Ch}}_{G,M} = 10.69$, $6.21$, $5.33$, and $4.48$ for w/o $\mathcal{F}$, w/o adaptive, w/o $E_s$, and the full model, respectively. 
Omitting $\mathcal{F}$ always leads to a significant error increase, and abandoning our adaptive training policy increases it by $39\%$. 
Fig.~\ref{fig:ablation} shows qualitative results  %
confirming the statistics over all sequences, \textit{i.e.,} the largest errors are present in the  colour-coded error maps when $\mathcal{F}$ is  disabled. 
The second most crucial component of $\boldsymbol{\phi}$-SfT is the adaptive training strategy. 
Note that $E_s$ helps when the structure deforms and significantly changes its shape in the re-projection. 
See our supplement for the complete report.

\subsection{Discussion and Future Directions} 
\begin{figure}[h]
\begin{center}
   \includegraphics[width=0.97\linewidth]{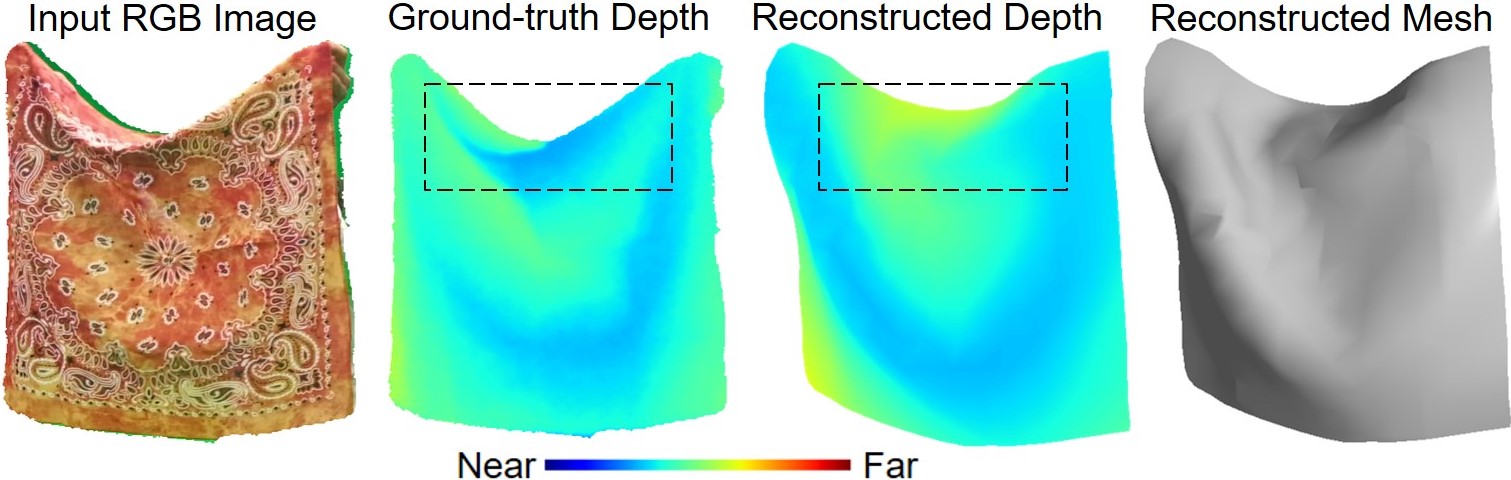} 
\end{center}
    \vspace{-8pt}
   \caption{
   Limitation: Due to the high runtime of the simulator, we use a lower-resolution mesh, which suffices to capture local folds but limits the reconstruction of very fine wrinkles (dashed area).
   }
\label{fig:limitations}
\end{figure}

We use PyTorch3D~\cite{ravi2020pytorch3d} for rendering; it only supports a single light source, which can 
differ from the illumination during the recording. 
There is, therefore, a systematic mismatch between the rendered reconstructions during training and the input images. 
Still, $\boldsymbol{\phi}$-SfT achieves high accuracy, as our  differentiable rendering loss is empirically robust to 
small noise and mismatches in colour and shading. 
The runtime of our approach is comparably high (\textit{i.e.,} twice as long as N-NRSfM \cite{Sidhu2020}). 
This is due to single-threading when resolving collisions which can be improved through parallelisation. 
Next, we use lower-resolution meshes for all sequences, with ${\sim}300$ vertices, as 
optimising model parameters with higher resolutions is not feasible. 
This limits the reconstruction accuracy as we cannot capture very fine wrinkles, as shown in Fig.~\ref{fig:limitations}. 
However, note that our photometric energy is defined densely over all pixels, instead of just the vertices, and thus uses the information in the high-resolution texture map. 
Therefore, the lower number of vertices does not impede  $\boldsymbol{\phi}$-SfT's ability to capture folds. 
Moreover, even more sophisticated physics simulators could be implemented, taking into account more physical laws (\eg, wind turbulence or electrostatic forces). 
Such requirements depend on downstream applications such as game and movie production or industrial quality control.

In this first work of its kind, we focus on investigating how a physics-based deformation model can be leveraged in classical SfT. 
Accurate inference of deformation forces and materials remains a  difficult problem that deserves more investigation in the future. 
Nonetheless, $\boldsymbol{\phi}$-SfT 
infers them well enough to enable intuitive editing; see the supplement. 
While we do not target complex objects or new  scenarios---such as the separate field of 
virtual garment and dress simulations \cite{White2007,Guo2021}---this can be an interesting future avenue.

\section{Conclusion} 
We introduced $\boldsymbol{\phi}$-SfT, a new optimisation-based SfT method that models deformations with a physical simulator and uses differentiable rendering to define a reprojection energy term over all pixels, exploiting texture information independent of the mesh resolution. 
Experiments on the new dataset demonstrate that our approach improves the reconstructions qualitatively and quantitatively by a significant margin over competing techniques of several method classes. 
Especially remarkable is $\boldsymbol{\phi}$-SfT's 
accuracy in folded surface regions. 
We believe that the proposed technique has a high potential for future research, and we hope to see more improvements on physically principled methods for monocular non-rigid 3D reconstruction. 

\noindent\textbf{Acknowledgement.} 
This work was supported by the ERC Consolidator Grant \textit{4DRepLy} (770784).

{\small
\bibliographystyle{ieee_fullname}
\bibliography{egbib}
}

\end{document}